\documentclass[10pt,twocolumn,letterpaper]{article}

\usepackage{times}
\usepackage{epsfig}
\usepackage{graphicx}
\usepackage{amsmath}
\usepackage{amssymb}
\usepackage[utf8]{inputenc} 
\usepackage[T1]{fontenc}    
\usepackage{url}            
\usepackage{booktabs}       
\usepackage{amsfonts}       
\usepackage{nicefrac}  
\usepackage{microtype} 
\usepackage{bm}
\usepackage{bbm}
\usepackage{times}
\usepackage{authblk}
\usepackage{epsfig}
\usepackage{xcolor,colortbl}
\usepackage{graphicx}
\usepackage{color}
\usepackage{stmaryrd}
\usepackage{algorithm}
\usepackage{algorithmic}

\begin{document}

\title{SEMEDA: Enhancing Segmentation Precision with Semantic Edge Aware Loss}

\author[1]{Yifu Chen}
\author[1]{Arnaud Dapogny}
\author[1, 2]{Matthieu Cord}

\affil[1]{\small LIP6, Sorbonne Universit\'e, 4 Place Jussieu, Paris, France}
\affil[2]{\small Valeo.ai, Paris, France}

\date{}
\maketitle

\begin{abstract}
  While nowadays deep neural networks achieve impressive performances on semantic segmentation tasks, they are usually trained by optimizing pixel-wise losses such as cross-entropy. As a result, the predictions outputted by such networks usually struggle to accurately capture the object boundaries and exhibit holes inside the objects. 
  
  In this paper, we propose a novel approach to  improve the structure of the predicted segmentation masks.  We introduce a novel semantic edge detection network, which allows to match the predicted and ground truth segmentation masks. This Semantic Edge-Aware strategy (SEMEDA) can be combined with any backbone deep network in an end-to-end training framework.
 Through thorough experimental validation on Pascal VOC 2012 and Cityscapes datasets, we show that the proposed SEMEDA approach enhances the structure of the predicted segmentation masks by enforcing sharp boundaries and avoiding discontinuities inside objects, improving the segmentation performance. In addition, our semantic edge-aware loss can be integrated into any popular segmentation network without requiring any additional annotation and with negligible computational load, as compared to standard pixel-wise cross-entropy loss.
\end{abstract}

\section{Introduction}

\begin{figure}[ht]
	\centering
	\includegraphics[width=\linewidth]{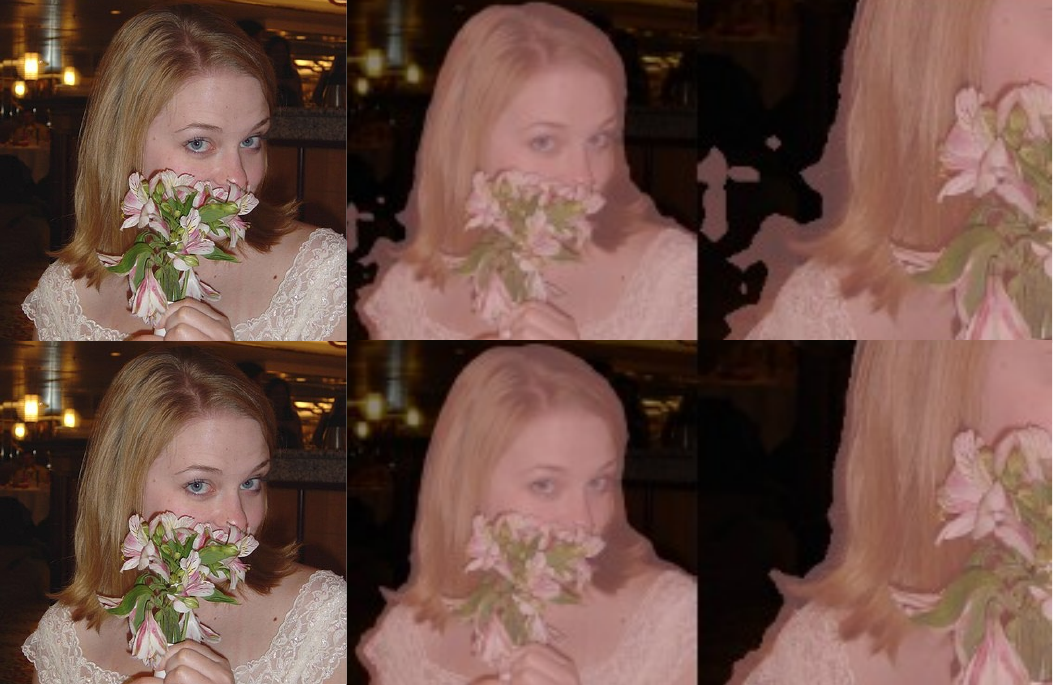}
	\caption{Left column: original image. Top row: prediction of a baseline segmentation model. Bottom row: prediction of the same model with the proposed edge-aware semantic loss. While the predictions of the baseline network do not align well with the shape of the objects, SEMEDA enforces structure on the segmentation masks.}
	\label{illustration}
\end{figure}

Semantic segmentation is one of the fundamental domains of computer vision, which aims at assigning a semantic label to each pixel of an image. Current state-of-the-art methods~\cite{LCChen2016,LCChen2017,Fu2017,LCChen2018,Zhao2017,Zhang2018,Lin2017} mainly rely on fully convolutional neural networks~\cite{Sermanet2014,Long2015} architectures, that are trained by optimizing a per-pixel loss between predictions and ground truth labels. However, as a result, the predicted maps are generally coarse and their boundaries do not align well with the shape of objects, as illustrated on Figure~\ref{illustration} (top row). In fact, the loss caused by discarding spatial information with down-sampling operations (\textit{e.g.} max-pooling or strided convolution) is one of the main reasons for such coarse prediction. Despite many improvements in the CNN architectures to avoid such pitfall, current networks still struggle to get segmentation results that accurately capture object boundaries.

A natural way to tackle this issue is to better integrate low-level information such as pixel affinity~\cite{Ke2018,Liu2017,Tu2018} and object boundaries~\cite{LChen2016DT,Bertasius2016,Marmanis2018} within high level features in order to output fine-grained predictions. Some works adopt probabilistic graphical models such as Dense CRF~\cite{Krahenbuhl2012} with efficient inference algorithm~\cite{Adams2010} while others design architectures that combine low level and high level features before final prediction~\cite{Long2015,Noh2015,Ronneberger2015,Badrinarayanan2017,Zhang2018exfuse}. Though these works are efficient at refining the segmentation result, a common assumption across almost all these approaches is that all label variables are learned with a per-pixel loss.

In this paper, we propose a  visual segmentation pipeline with a novel learning scheme that leverages semantic boundary information in the loss function. Departing from a classical multi-head architecture integrating an edge map output loss, we introduce an additional deep net over the segmentation mask to produce a semantic edge map. With this dedicated net, we derive an edge-aware loss that leverages spatial distribution of the predicted pixel labels. Our method is called SEMEDA for SEMantic EDge-Aware strategy.

We experimentally demonstrate the effectiveness of the proposed model on PASCAL VOC 2012 and Cityscapes data sets. The contributions of this paper are threefold:

\begin{itemize}
    \item We propose a novel way to integrate boundary information for semantic segmentation tasks that consists in matching the prediction and ground truth masks within the embeddings of a semantic edge detection network.
    \item Our experimental results on PASCAL VOC 2012 and Cityscapes datasets show that our approaches lead to substantial improvements. In addition, this method can be applied to enhance any deep network trained for semantic segmentation tasks.
    \item Furthermore, we benchmark a number of approaches for enforcing structure using boundary information on the predictions of a semantic segmentation network, highlighting the caveats of popular approaches such as multi-task and direct boundary matching methods.
\end{itemize}

The rest of the paper is organized as follows. Section 2 presents a brief review of related works. Section 3 presents our segmentation strategy using the extra deep net and our semantic edge-aware loss. In Section 4, many experiments on semantic segmentation are reported.

\section{Related Work}

Methods based on Fully Convolutional Networks~\cite{Sermanet2014,Long2015} have showed significant improvement on several segmentation benchmarks~\cite{everingham2015pascal,Cordts2016, Zhou2017, Caesar2018}. The early approaches can predict coarse masks but struggle to output accurate object boundaries. This is mainly because that detailed spatial information is lost during down sampling operations. A big improvement has been obtained by using "atrous" convolution~\cite{LChen2015} to increase the spatial resolution of the segmentation masks.

Many strategies have been proposed to exploit spatial low-level information to refine segmentation result. 
The first idea that incorporates low level information is to combine the prediction with  probabilistic graphical models~\cite{LiuZ2015,LChen2015,Zheng2015,Ke2018,Bertasius2016,Bell2015,Vemulapalli2016, Chandra2016,Chandra2017,Lin2016}. Prior work proposed to use post-processing methods such as CRF~\cite{LChen2015} and Domain Transform~\cite{LChen2016DT} to refine the output of the segmentation network. However, these methods add new parameters which are difficult to tune and integrate into the original network architecture. Authors in~\cite{Zheng2015} formulate mean-field approximate inference for CRF with Gaussian pairwise potentials as recurrent neural networks and then can be trained end-to-end with segmentation network. These works have demonstrated promising results on the refinement of the predicted segmentation masks.

Another commonly used strategy is to employ encoder-decoder architectures~\cite{Long2015,Noh2015,Ronneberger2015,Badrinarayanan2017,Lin2017,Pohlen2017,Peng2017,Islam2017,Wojna2017,Fu2017,Zhang2018exfuse}. These models generally consist of two parts: an encoder module reduces the feature map resolution and capture semantic level information, and a decoder module that gradually recovers the spatial information. The idea is that the low-level information are encoded in hidden layers of the encoder, and thus can be recovered by reusing these features. \cite{Long2015, Noh2015} use deconvolution~\cite{Zeiler2011} to learn the upsampling while~\cite{Badrinarayanan2017} reuses the pooling indices from the encoder and learn extra convolutional layers. U-Net~\cite{Ronneberger2015} adds skip connections from the encoder features to the corresponding decoder activations. This idea is adopted by many works~\cite{Pohlen2017, Lin2017,LCChen2018, Peng2017, Islam2017, Wojna2017, Fu2017,Zhang2018exfuse}, and have demonstrated the effectiveness of models based on encoder-decoder structures on several semantic segmentation benchmarks.

Though these works are efficient at refining the segmentation result, a common assumption across almost all these approaches is that all label variables are learned with a per-pixel loss. This loss, which treats each pixels equally, aims to minimize the average accuracy of all pixels. This loss is often a poor indicator of the quality of the segmentation when there is a class imbalance problem since it does not take recall into account. Moreover, the evaluation metric used in tests is the Jaccard index, also called the intersection-over-union (IoU) score. This measure is not differentiable and thus cannot be used for training segmentation networks. Some approaches employed approximated Jaccard index loss~\cite{Berman2018,Nagendar2018} or generalized dice loss~\cite{Sudre2017} 
 and slightly improve the mean IoU score. However, this measurement does not accurately describe the structure of the segmentation mask, as argued in~\cite{Csurka2013}.

We are interested in finding a suitable objective function which is able to incorporate the structure of the segmentation mask during training. \cite{luc2016semantic, Hung2018} propose to add an adversarial loss which enforces the prediction to conform high-level structure such as shape of objects with ground truth segmentation masks. Unlike instance segmentation, the shape of semantic segments is not well defined in semantic segmentation (e.g, a ground truth label does not separate two object of the same class if they are overlapped). In addition to partial occlusion, object scaling, camera viewpoint change and small size of segmentation data set, the latent "structural" distribution of the masks is extremely complicated and difficult to be learned by a discriminator network. 

Instead of learning this structural distribution, we propose a semantic edge-aware loss which directly constraints the predicted segmentation masks to have the same semantic edge as its ground truth counterpart. 

\begin{figure*}[ht]
	\centering
	\includegraphics[width=\linewidth]{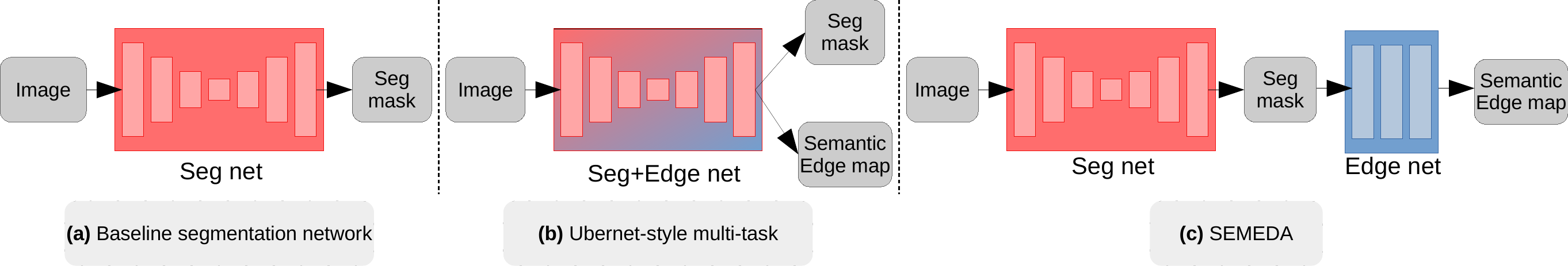}
	\caption{Segmentation deep net architectures with edge-aware loss strategies. The baseline strategy (a) is based on a backbone deep architecture (Seg net) providing a segmentation mask from the input image. The learning loss is defined on the output mask. To get sharper segmentation masks, (b) presents the basic extension that adds an explicit loss working with semantic edges.  semantic segmentation with semantic edge detection. Our method, SEMEDA, presented in (c) works differently, as well in term of deep architecture extension (Section~\ref{archi}), as for the corresponding loss framework (Section~\ref{learning}).}
	\label{archis}
\end{figure*}

\section{SEMEDA: Semantic Edge-Aware Loss}

\subsection{Motivations}
In this section, we present our segmentation strategy. Our goal is to use the semantic edge detection to enhance segmentation task. Our pipeline uses a backbone segmentation deep network, and we add a new dedicated branch to carry out our semantic edge-aware loss.  
Figure~\ref{archis} illustrates the basic idea behind our approach. In Figure~\ref{archis}-(a), a baseline deep network is trained to map an input image to a segmentation mask. Figure~\ref{archis}-(b) describes a naive Ubernet-style~\cite{kokkinos2017ubernet} approach that adds a new head to the segmentation network, that shall infer the semantic edge maps from the image. The segmentation and edge detection networks share a common backbone, hoping that the edge detection task helps refining the semantic segmentation. However, it should be noted that such architecture does not impose any constraint on the segmentation masks. Even worse, there is no guarantee that the edge detection task does not hinder the segmentation task instead of driving it. 

Conversely, we propose an architecture where semantic edges are inferred from the segmentation mask using a new network. It is referred as Edge net in the Figure~\ref{archis}-(c) that illustrates our proposed architecture.

This way, the segmentation mask can be structurally constrained by aligning the predicted and ground truth semantic edges during training. This strategy,  called SEMEDA (in reference to the  semantic edge-aware loss), can be inserted into any state-of-the-art segmentation network architectures. The whole model is trainable end-to-end without additional annotations and adds negligible computational overhead.

\subsection{Edge detection network}
\label{archi}
\begin{figure*}[ht]
	\centering
	\includegraphics[width=\linewidth]{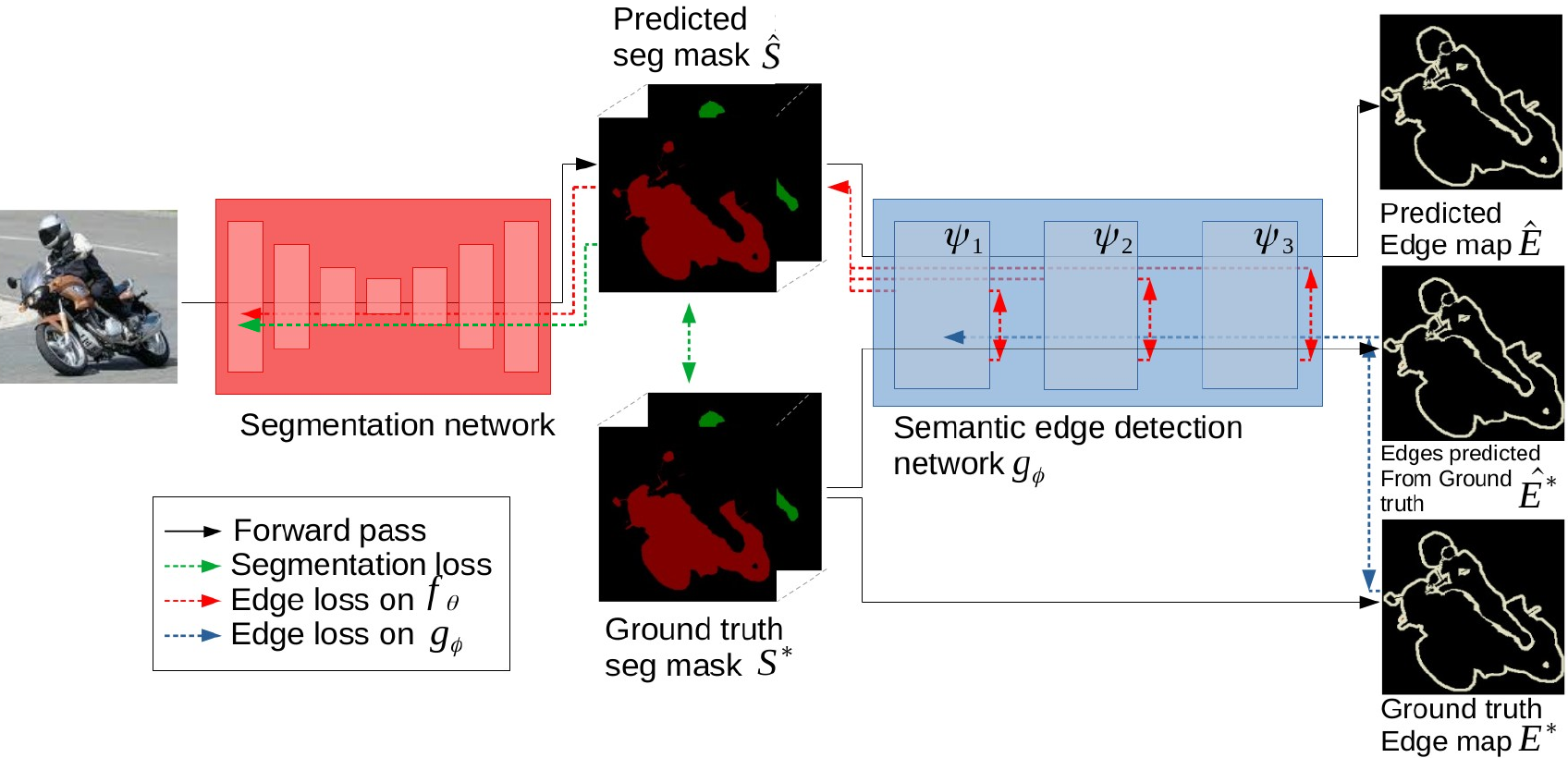}
	\caption{Overview of our semantic edge-aware loss for semantic segmentation. Both the ground truth segmentation mask, and the mask predicted by a segmentation network $f_{\theta}$ are fed into an edge detection network $g_{\phi}(S)$, that is trained by minimizing (blue dashed lines) pixel-wise cross-entropy between a ground truth edge map $E^*$ and an edge map predicted from the ground truth mask $\hat E^*$. The embeddings of this network are then matched to train the segmentation network (red dashed lines).}
	\label{main}
\end{figure*}

In order to instantiate our SEMEDA strategy depicted in Figure~\ref{archis}-(c), we need to define the Edge net, the additional losses and learning schemes.
Our proposition for this architecture is depicted in Figure~\ref{main} and is composed of two modules: 
\begin{itemize}
    \item $f_{\theta}$ a semantic segmentation network parameterized by weights $\theta$ (Seg net). This network can be any popular network, it maps an RGB image $I \in \mathbf{R}^{H*W*3}$ into a segmentation mask $\hat{S}$:
    \begin{equation}\hat{S} = f_{\theta}(I) \in \mathbf[0,1]^{H*W*C}\end{equation}
    where $H$ (resp. $W$) is the height (resp. weight) of the input image and $C$ is the total number of semantic classes (including background).
    
    \item $g_{\phi}$ a semantic edge detection network (Edge net). 
    This network maps a segmentation mask $S \in \mathbf[0,1]^{H*W*C}$ into a binary edge map $\hat{E}$:
    \begin{equation}\hat{E}  = g_{\phi}(S) \in \mathbf[0,1]^{H*W*2}.\end{equation}
    Because the task of detecting edges from the segmentation mask is a rather straightforward one, we use a simple convolutional network composed of $L=3$ layers with $16 \rightarrow 32 \rightarrow 2$ channels and ReLU activation (except for the last layer, which has SoftMax activation) to model the function $g_{\phi}$. In what follows, we respectively denote $\psi^1$,$\psi^2$,$\psi^3$ the embeddings of these convolutional layers.   
\end{itemize}
 
Figure~\ref{main} illustrates the whole architecture as well as the forward and backward flow through the semantic segmentation network as well as the edge detection network, at train time. 
 
\subsection{Structure learning through edge detection}
\label{learning}
The canonical loss for training deep networks for semantic segmentation is the Per-Pixel Cross Entropy (PPCE) loss $\mathcal L^{PPCE}$. This loss function measures the difference between the predicted label mask $\hat{S}$ and the ground truth mask $S^{*}$ by:

\begin{equation}\label{ppce}
\mathcal L^{PPCE}(\hat{S}, S^{*}) = - \sum \limits_{i=1}^H \sum \limits_{j=1}^W \sum_{c=1}^{C} S^{c*}_{i,j} \log(\hat{S}_{i,j}^{c})\,.
\end{equation}

A segmentation model that does not consider edge detection, such as the one described on Figure~\ref{archis}-(a), is usually trained upon optimizing PPCE. PPCE treats each pixel equally and does not take the structure of the segmentation mask into account. Now, given a ground truth segmentation mask $S^{*}$, we can obtain a corresponding semantic edge map $E^{*}$ in a straightforward fashion: to do so, we define edge pixels as pixels that do not have 8 identical labeled neighbour pixels, and other pixels as non-edge. These edge maps are calculated one time beforehand and no further computation is needed afterwards. Then, the edge detection network $g_{\phi}$ is trained by computing:
\begin{equation}\label{edgedetectiontrain}
\phi^{*}=argmin_{\phi} \mathcal L^{PPCE}(\hat{E}^{*}, E^{*})
\end{equation}

Where $\hat{E}^{*}=g_{\phi}(S^{*})$ Now, given these ground truth semantic edge maps, and the trained semantic edge detection network $g_{\phi^*}$, we can address the structure issue by directly constraining the edges $\hat{E}$ of the predicted segmentation mask $\hat{S}$ to look like the edges $E^{*}$ of the ground truth segmentation mask $S^{*}$, \textit{e.g.} by applying PPCE between $\hat{E}$ and $E^{*}$ and optimizing the loss:
\begin{equation}\label{ppceonedges}
\theta^*=argmin_{\theta} \mathcal L^{PPCE}(\hat{S}, S^{*}) + \lambda_1 \mathcal L^{PPCE}(\hat{E}, E^{*})
\end{equation}

w.r.t. segmentation network $f_{\theta}$ only (Figure~\ref{archis}-(c)).

In what follows, we refer to this method as PPCE on edges. However, as pointed out in~\cite{johnson2016perceptual}, the embeddings of deep networks capture more abstract information as we stack convolutional layers as well as non-linearities. Hence, PPCE on edges is not the most optimal way of matching the two segmentation masks using the edge detection network. Also note that the basic Ubernet-style multi-task formulation (\cite{kokkinos2017ubernet}-Figure~\ref{archis}-(b)) optimizes that same loss, however the gradient backpropagated through the segmentation head does not take the edge constraint into account, which is why this architecture does not enforce structural information into the predictions. 

In the frame of style transfer and image synthesis, the authors of~\cite{johnson2016perceptual} obtained impressive results by defining high-level perceptual losses that involve a fixed pre-trained network, such as ImageNet-pretrained VGG-19~\cite{Simonyan15} network. The idea of this method is to measure the semantic difference between two images as the difference of feature representation as computed by the fixed network. However, this idea cannot be translated to semantic segmentation in a straightforward fashion, as natural images and segmentation masks have different number of channels and belong to wildly different distributions. Nevertheless, inspired by this work, we propose to define our edge-aware loss as:
\begin{equation}\label{edgelosseq}
\mathcal L_{\text{SEMEDA}}^{g_{\phi^*}}(\hat{S},S^{*}) = \sum \limits_{i=1}^H \sum \limits_{j=1}^W \sum_{l=1}^{L} \lambda_{l} |\psi^{l}_{i,j}(\hat{S}) - \psi^{l}_{i,j}(S^{*})|
\end{equation}
where $\psi^{l}_{i,j}$ represents the pixel $(i,j)$ of the embedding at depth $l$ of the semantic edge detection network $g_{\phi^*}$ ($\psi^{l}_{i,j}$ is thus a $c$-dimensional vector, where c is the number of channels in layer $l$) and $\lambda_{l}$ is a hyper parameter representing the importance of this layer in the total loss.

The proposed semantic edge detection network $g_{\phi}$ is thus trained to minimize PPCE between edge maps $\hat E^{*}$ prediced from the ground truth segmentation masks $S^{*}$ and ground truth edge maps $E^{*}$, as indicated by the blue arrows in Figure~\ref{main}. However, $S^{*}$ is one hot-encoded while $\hat{S}$ (the prediction of the segmentation network) is the output of a softmax layer. In order to mimic the distribution of $\hat{S}$, we add Gaussian noise to $S^{*}$ and then apply a softmax function before passing through the edge detection network. Once this network is trained, we fix all parameters $\phi^*$ of the network $g_{\phi^*}$ and use it to train the segmentation network $f_{\theta}$ upon loss $L_{\text{SEMEDA}}^{g}$ red arrows in Figure~\ref{main}). Our final loss is a combination of the PPCE loss on the segmentation masks and the proposed edge-aware loss:

\begin{equation}\label{edgelosstot}
\mathcal L_{tot}^{g_\phi^*}(\hat{S},S^{*}) = \mathcal L^{PPCE}(\hat{S}, S^{*}) + \mathcal L_{\text{SEMEDA}}^{g_\phi^*}(\hat{S},S^{*})\,.
\end{equation}

The steps for training a segmentation network with SEMEDA are summarized in Algorithm~\ref{algo}. In what follows, we compare several configurations for training segmentation networks with edge-aware loss, and show the effectiveness of the proposed approach.

\begin{algorithm}
\caption{Train a segmentation network with SEMEDA}
\label{algo}
\begin{algorithmic}
\REQUIRE 
\STATE \makebox[2em][l]{$I$} RGB Images
\STATE \makebox[2em][l]{$S^{*}$} Ground truth segmentation masks 
\ENSURE 
\STATE \makebox[2em][l]{$\theta^*$} Parameters of the segmentation net
\STATE
\FORALL{masks $S^*_k$,$k=1,...,K$ in a batch}
\STATE Generate ground truth edge map $E^{*}_k$ from $S^{*}_k$ by
\STATE examining neighbouring pixels labels
\STATE $\hat E_k ^ * = g_{\phi}(S^{*}_k)$
\STATE $\phi\leftarrow\phi-\frac{1}{K}\frac{\partial}{\partial \phi} \mathcal L^{PPCE}(\hat E_k^ *,E^{*}_k)$
\ENDFOR
\STATE \emph{// edge net is trained with parameters $\phi^*$}
\FORALL{labelled images $I_k$ in in a batch}
\STATE $\hat S_k  = f_{\theta}(I_k)$
\STATE $\theta\leftarrow\theta-\frac{1}{K}\frac{\partial}{\partial \theta} \mathcal L_{tot}^{g_\phi^*}(\hat{S}_k,S^{*}_k)$
\ENDFOR
\end{algorithmic}
\end{algorithm}

\section{Experiments}

We present in this section our experiments to validate our SEMEDA approach.
We use two challenging segmentation datasets, and provide first detailed analysis of all the parameters and the different options for learning our models. Then, quantitative and qualitative results are provided.

\subsection{Datasets}
We conduct our experiments on the Pascal VOC 2012 database~\cite{everingham2015pascal}, which contains 20 foreground object classes as well as one background class.  We train our models on the augmented version of the dataset~\cite{hariharan2011semantic} which contains 10.582 training images and report results on the validation partition which contains 1449 images.

We also evaluate our models on the \textbf{Cityscapes} database, which contains 2975 images annotated with quality pixel-wise annotations of 18 object classes and one background class for training. We report results on the validation set, which contains 500 images. For memory reasons, we downsample images to half resolution, \textit{i.e.} $1024 \times 512$ for both training and evaluation.

\subsection{Implementation details}
We employe Deeplab-v2~\cite{LChendeeplabv2} as our backbone segmentation network with a minor modification: we add a bilinear upsampling layer before the softmax layer to output a mask of the same size as the input image. The edge detection network is pre-trained as mentioned in Section 3, and is fixed for defining edge-aware loss. In order to keep the runtimes and memory footprint reasonable, we train our models by feeding to the networks 321 x 321 random crops for both datasets without multi-scale inputs. We adopt random scaling and random mirror strategies before cropping. Since we use mini-batch of 6 images, we fix parameters in batch norm layers and initial learning rate of $5 \cdot 10^{-4}$ ($5 \cdot 10^{-3}$ for final classifier layer). As is classically done in the literature, we report the mean intersection over union (mIoU) metric over all the classes as our evaluation metric.

\subsection{Ablation study}\label{aablation}

\begin{table}[]
\caption{Comparison of different models with different values of $\lambda_{l}$ in Equation \eqref{edgelosseq} ($\%$ mIoU) on Cityscapes. For Ubernet-style multi-task model and PPCE on edges, there is only one hyperparameter which is the coefficient of the edge term $\lambda_1$}
\label{ablation}
\begin{tabular}{|l|c|c|c|r|}
\hline
Loss                & $\lambda_1$& $\lambda_2$ & $\lambda_3$ & mIoU \\
\hline
PPCE \textit{(Fig~\ref{archis}-(a)-Eq \eqref{ppce})}            & -       & -       & -                        & 64.1   \\
\hline
Ubernet-style multi-task & 1       & -       &  -                  & 64.6   \\
\textit{(Fig~\ref{archis}-(b)-Eq \eqref{ppceonedges})}                       & 0.5       & -       &    -                & 64.3     \\
                      & 5       & -       &    -                 & 64.3   \\
\hline
PPCE on edges         & 1       & 0       & -                    & 64.1    \\
\textit{(Fig~\ref{archis}-(c)-Eq \eqref{ppceonedges})}                     & 5       & 0       &  -                 & 64.7   \\
\hline
SEMEDA (after ReLU) & 1       & 0       & 0                  & 65.5    \\
\textit{(Fig~\ref{archis}-(c)-Eq \eqref{edgelosstot})}       & 0     & 0.5       & 0                       & 65.9    \\
\hline
SEMEDA  (before ReLU) & 1      & 0       & 0             & 66.0    \\
\textit{(Fig~\ref{archis}-(c)-Eq \eqref{edgelosstot})}       & 0.5     & 0       & 0               & 65.6    \\
                      & 0       & 1       & 0                      & \textbf{66.6} \\
                      & 0       & 0       & 1                   & 65.6 \\
                      & 1       & 0.5       & 0.25                      & \textbf{66.5} \\
                      & 0.25       & 0.5       & 1                     & 66.3 \\
\hline  
\end{tabular}
\end{table}

Table~\ref{ablation} shows a comparison between different losses applied on Deeplab V2 model pre-trained on ImageNet. As echoed in~\cite{kokkinos2017ubernet}, using edge information to help the semantic segmentation task in a naive multi-task way does not significantly help the semantic segmentation task, whatever value the hyperparameter $\lambda_3$ is set at. This is due to the face that adding an additional head to predict edges do little to constrain and add structure to the segmentation masks.

 Furthermore, training a small CNN for semantic edge detection and optimize PPCE loss directly between the edge maps obtained from the predicted and ground truth masks does not significantly help the segmentation task either. In contrast, matching the embeddings of such network in terms of $\mathcal{L}_1$ or $\mathcal{L}_2$ loss provides higher mIoU scores for all the tested configurations. This result is similar to the use of perceptual losses~\cite{johnson2016perceptual} for natural images, where only the first layers of a pre-trained network such as VGG~\cite{Simonyan15} are used to match low-level statistics (\textit{e.g.} oriented gradients) between pairs of images. 

Similarly, our edge-aware loss allows the network to more efficiently capture the structure of the segmentation masks compared to PPCE loss. The best configurations are obtained with either:
\begin{itemize}
    \item $\lambda_1=0, \lambda_2=1, \lambda_3=0$, or
    \item $\lambda_1=1, \lambda_2=0,5, \lambda_3=0.25$.
\end{itemize}
  
  Interestingly, matching the embeddings of the edge detection network before ReLU activations is better than matching them after ReLU, as in the latter case we simply discard some information in the loss term. Nevertheless, for all the configurations, the proposed edge-aware loss provides a steady boost in accuracy. It should be emphasized that this enhancement virtually comes for free and could be applied to any state-of-the-art segmentation network.

\subsection{Quantitative evaluation}

We use DeepLab v2 network as a backbone architecture to quantitatively evaluate the improvement when applying our SEMEDA learning strategy.

Table~\ref{ablation2} summarizes results on VOC2012 and Cistyscapes databases, obtained with Imagenet and MSCoco-pretrained models. In either case, in all tested configurations, the proposed edge-aware loss allows to substantially increase the overall accuracy of the models, by enforcing structure and avoiding discontinuities inside the segmented objects and refining the inter-class boundary regions, whereas a naive Ubernet-style multi-task approach does not substantially increase the performance.

To more precisely assess the performance of edge-aware loss, we propose to evaluate the performance of our method on boundary/non-boundary trimaps, as in~\cite{LChen2016DT,Csurka2013}: at test time, we divide the pixels in two subsets, whether they belong to a boundary or non-boundary region, as indicated by the semantic edge maps generated from the ground truth segmentation masks. To do so, we vary the width of a band centered on the boundary and count as positive all the pixels in the region defined by this band, negative otherwise: thus, the more the width increases, the less precise the boundary definition is. 

\begin{table}[]
\caption{\% mIoU on VOC2012 and Cityscapes database with different pre-training strategies.}
\label{ablation2}
\begin{tabular}{|l|r|r|}
\hline        
& VOC & Cityscapes  \\
\hline
\multicolumn{3}{|l|}{Pre-trained ImageNet }   \\
\hline
deepLab v2                           & 0.729           & 0.641  \\
deepLab v2 + Edge Ubernet-style & 0.729               & 0.646  \\
deepLab v2 + SEMEDA            & \textbf{0.742}           & \textbf{0.666}  \\
\hline
\multicolumn{3}{|l|}{Pre-trained ImageNet+MsCoco }   \\
\hline
deepLab v2                           & 0.753           & 0.658                  \\
deepLab v2 + Edge Ubernet-style & 0.756               & 0.664           \\
deepLab v2 + SEMEDA              & \textbf{0.765}           & \textbf{0.678}             \\
\hline
\end{tabular}
\end{table}

\begin{figure*}[t]
	\centering
	\includegraphics[width=\linewidth]{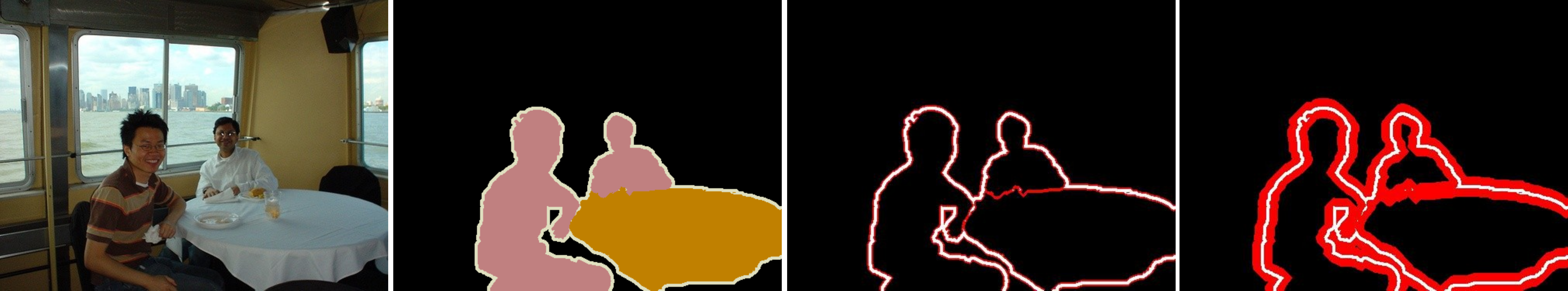}
	\caption{Left to right: original image, ground truth segmentation and two boundary/non-boundary trimaps: one with 1 pixel width and the other with 10 pixels width. The red areas represent the boundary regions while black areas represent non-boundary regions. White areas are 'void' pixels.}
	\label{visu_trimaps}
\end{figure*}

\begin{figure*}[ht]
	\centering
	\includegraphics[width=\linewidth]{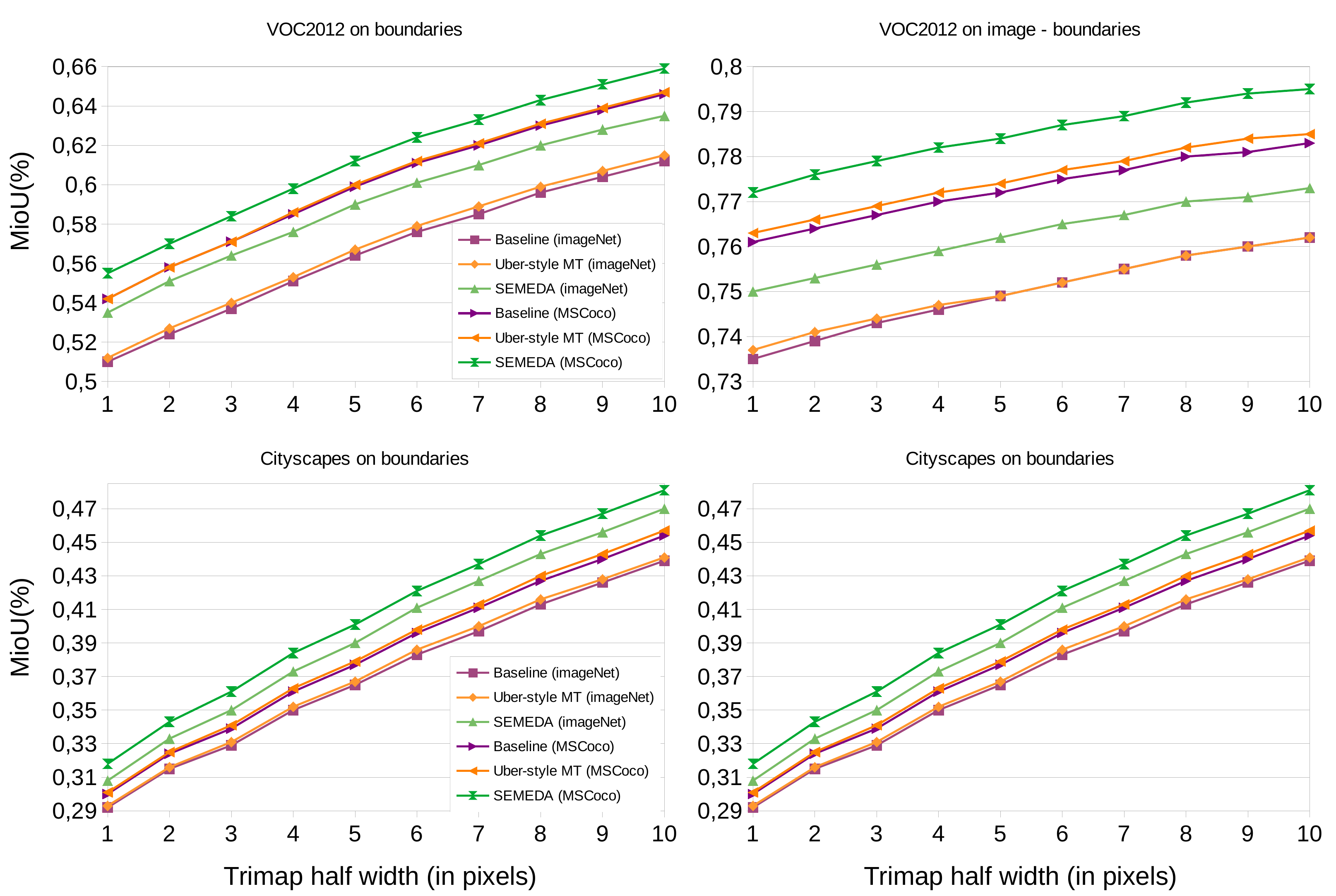}
	\caption{mIoU for different models on boundary and non-boundary regions on VOC2012 and CityScapes datasets. For both the models pretrained on imageNet and MSCoco, Edgle loss substantially enhances the performance on both regions.}
	\label{trimaps}
\end{figure*}

This process is illustrated on Figure~\ref{visu_trimaps} with ImageNet and MSCoco pretrained models. The proposed edge-aware loss allows to significantly enhance the mIoU on the boundary regions on both databases, meanwhile the naive Ubernet-like~\cite{kokkinos2017ubernet} method lies closer to the baseline performance. Particularly for strict boundaries (trimap width $1,2$), the mIoU improvement is $4.4\,$pts on Pascal VOC and $3.7\,$pts on Cityscapes, which is considerable. On non-boundary regions, the improvement is also very significant on both datasets, as Imagenet-pretrained models with edge-aware loss are better than MSCoco-pretrained baseline models. This is due to the fact that edge-aware loss strongly penalizes the presence of holes or discontinuities in the internal structure of the predicted objects (which are tagged as non-boundary on the ground truth markups). Thus, edge-aware loss allows to better capture the structure of objects, as well as to refine the boundaries between different classes in the segmentation masks.

\subsection{Qualitative results}

\begin{figure*}[p]
	\centering
	\includegraphics[width=0.9\linewidth]{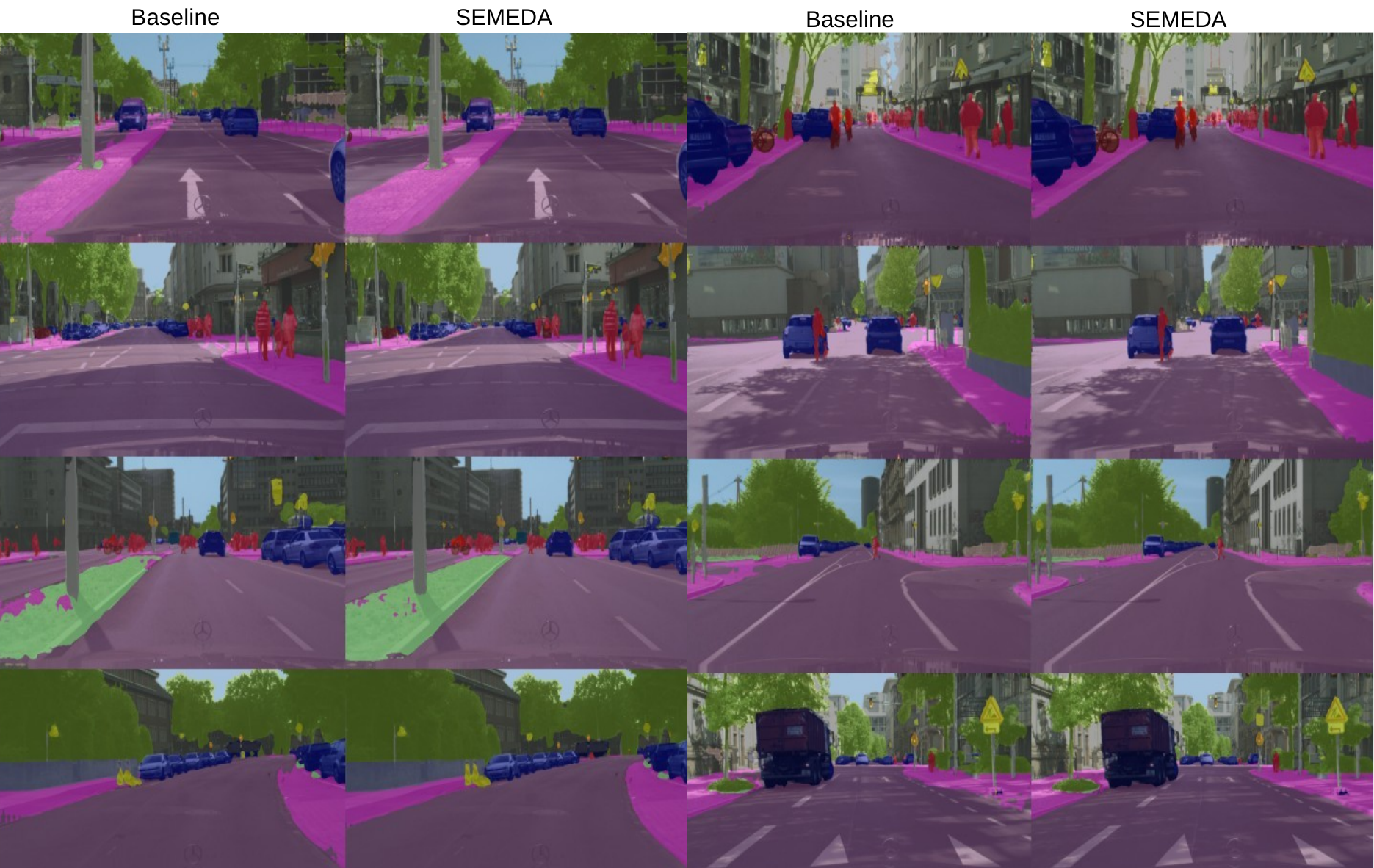}
	\caption{Examples of predicted segmentation masks on Cityscapes val set.}
	\label{visu_city}
\end{figure*}

\begin{figure*}[p]
	\centering
	\includegraphics[width=0.9\linewidth]{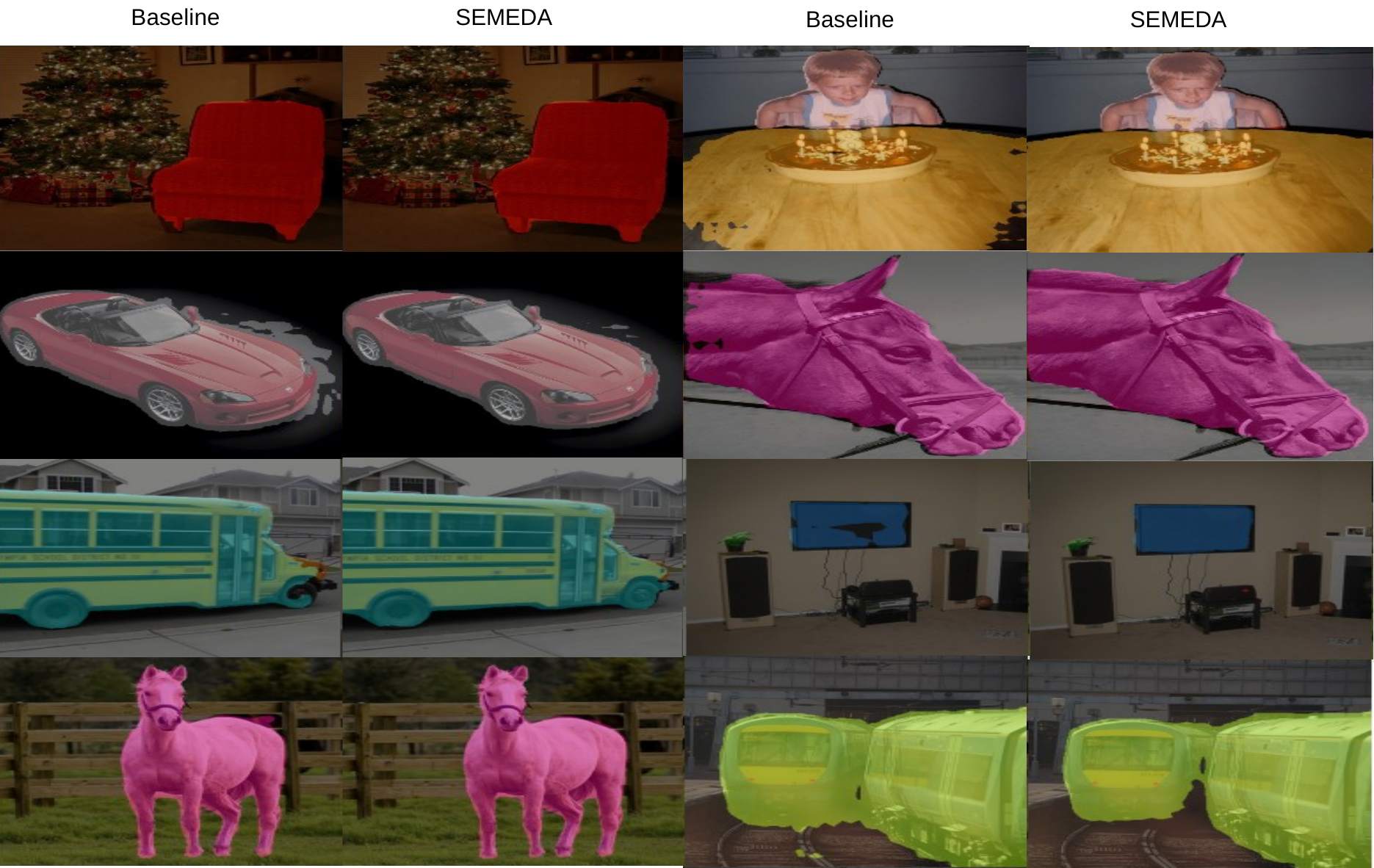}
	\caption{Examples of predicted segmentation masks on Pascal VOC 2012 val set.}
	\label{visu_voc1}
\end{figure*}

Figures~\ref{visu_city} and~\ref{visu_voc1} shows results outputted by baseline and edge-aware loss-pretrained models trained on Pascal VOC 2012 and Cityscapes datasets, respectively. For each image, the segmentation mask provided by the network is overlayed with the input image. Overall, as stated above, edge-aware loss allows to better capture the structure of the segmented objects, by putting more emphasis on the inter-class boundaries, as well as to avoid discontinuities (\textit{e.g.} holes) inside the object structures. Notice, for instance, how fine-grained elements such as traffic signs, tree leaves or people shapes are better captured with edge-aware loss on Cityscapes, and how well the segmentation fits the objects on Pascal VOC 2012.

\section{Conclusion}

In this paper, we proposed a new learning scheme for visual segmentation framework.
Our approach, SEMEDA, leverages a semantic edge-aware loss for implicitly integrating structural information into segmentation predictions. It consists in training a semantic edge detection network to map segmentation masks to the corresponding edge maps. The predictions outputted by the segmentation network can then be optimizing in the embedding space of the semantic edge detection network, similarly to what is done with perceptual losses. 

We showed that our edge-aware loss framework significantly improves the overall performance of semantic segmentation networks by enforcing inter-class boundary structure as well as avoiding holes in the segmented objects. In addition, edge-aware loss does not require any additional annotation and very small computational burden, thus can be applied to improve the performance of any semantic segmentation network.

{\small
\bibliographystyle{ieee}

}

\end{document}